# Logarithmic Time Parallel Bayesian Inference

**David M. Pennock**
University of Michigan AI Laboratory
1101 Beal Avenue
Ann Arbor, MI 48109-2110 USA
dpennock@umich.edu

## Abstract

I present a parallel algorithm for exact probabilistic inference in Bayesian networks. For polytree networks with $n$ variables, the worst-case time complexity is $O(\log n)$ on a CREW PRAM (concurrent-read, exclusive-write parallel random-access machine) with $n$ processors, for any constant number of evidence variables. For arbitrary networks, the time complexity is $O(r^{3w} \log n)$ for $n$ processors, or $O(w \log n)$ for $r^{3w} n$ processors, where $r$ is the maximum range of any variable, and $w$ is the induced width (the maximum clique size), after moralizing and triangulating the network.

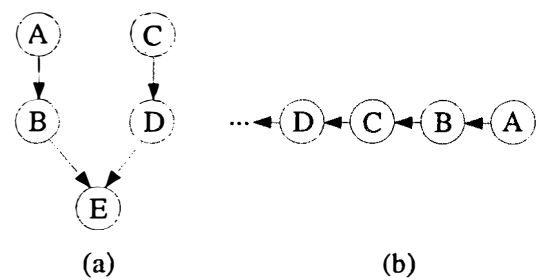

Figure 1: Comparing the Opportunities for Parallelism in Two Bayesian Networks. Network (a) seems to have more "natural" parallelism.

## 1 INTRODUCTION

Two key breakthroughs make representation of and reasoning with probabilities practical, and have led to a proliferation of related research within the artificial intelligence community. *Bayesian networks* exploit conditional independence to represent joint probability distributions compactly, and associated *inference algorithms* evaluate arbitrary conditional probabilities implied by the network representation (Neapolitan 1990; Pearl 1988). Exact inference is known to be NP-hard (more specifically, #P-complete) (Cooper 1990), and even approximate inference is NP-hard (Dagum and Luby 1993). Nonetheless, clever algorithms exploit network topology to make exact probabilistic reasoning practical for a wide variety of real-world applications. Although no algorithm, including any parallel algorithm with a polynomial number of processors, can avoid worst-case exponential time complexity (unless $P = NP$), polynomial time speed-ups are still of great interest, to allow evaluation of larger, more densely connected Bayesian networks.

Some researchers have proposed parallel algorithms for Bayesian inference (D'Ambrosio 1992; Diez and Mira 1994; Kozlov 1996; Kozlov and Singh 1994; Shachter, Andersen, and Szolovits 1994). In fact, Pearl's original algorithm for inference in singly connected networks was conceived as a distributed algorithm, where each variable could be associated with a separate processor, passing and receiving messages to and from only local, neighboring processors (Neapolitan 1990; Pearl 1988). The popular *junction tree* algorithm for multiply connected networks (Jensen, Lauritzen, and Olesen 1990; Lauritzen and Spiegelhalter 1988; Neapolitan 1990; Spiegelhalter, Dawid, Lauritzen, and Cowell 1993) can be considered a distributed algorithm in the same sense, with one processor per clique. However, to my knowledge, there are no derivations of improved worst-case complexity bounds for parallel inference, mainly because previous attempts at parallelization only exploit concurrency in the form of independent computations. For example, consider the two networks in Figure 1. In both (a) and (b), the two computations $\Pr(B = b_1) \leftarrow \sum_A \Pr(B = b_1|A)\Pr(A)$ and $\Pr(B = b_2) \leftarrow \sum_A \Pr(B = b_2|A)\Pr(A)$ are independent and easily parallelized. D'Ambrosio (1992) calls this a *conformal product* parallelization, while Kozlov and Singh (1996, 1994) call it a *within-clique* parallelization. In network (a), the computations $\Pr(B) \leftarrow \sum_A \Pr(B|A)\Pr(A)$ and $\Pr(D) \leftarrow \sum_C \Pr(D|C)\Pr(C)$ are also independent; this has been called *topological* parallelism. However in the chain network (b), the computa-

tion $\Pr(D) \leftarrow \sum_C \Pr(D|C) \Pr(C)$ must wait for the computation of $\Pr(C)$, which must in turn wait for the computation of $\Pr(B)$, etc. Because of the possibility of such chains, any implementation with one processor per variable (or one per clique), that only exploits obvious independent computations, will have the same worst-case running time as a serial algorithm. This paper describes a parallelization with improved time complexity, regardless of the network topology. To propagate probabilities in chains, it employs a procedure similar to *pointer jumping*, a standard "trick" used in parallel algorithms (Cormen, Leiserson, and Rivest 1992; Gibbons and Spirakis 1993; JáJá 1992). Polytrees are handled with a variant of *parallel tree contraction* (Abrahamson, Dadoun, Kirkpatrick, and Przyttycka 1989; JáJá 1992; Miller and Reif 1985). Evidence propagation is achieved with a parallel version of Shachter's arc reversal and evidence absorption (Shachter 1988; Shachter 1990) The computational model employed is the CREW PRAM, or the concurrent-read, exclusive-write parallel random-access machine; this model assumes a shared, global memory that processors can read from, but not write to, simultaneously (Cormen, Leiserson, and Rivest 1992; Gibbons and Spirakis 1993; JáJá 1992). The algorithm is termed PHISHFOOD, for **P**arallel algorit**H**m for **I**nference in Baye**S**ian networks wit**H** a (really) **FOO**rce**D** acronym. For **M**ultiply-**C**onnected networks, the name McPHISHFOOD is tastelessly coined.

The next section reviews Bayesian networks and probabilistic inference, and the necessary notation and terminology. For clarity of exposition, the PHISHFOOD algorithm is introduced in four stages, each successively more general. Section 3 presents a key parallelization component via a simple example, describing how to find all marginal probabilities in a chain network. Sections 4, 5, and 6 generalize the algorithm to find conditional probabilities in tree, polytree, and arbitrary networks, respectively. In all cases, for any constant number of evidence variables, the running time is $O(r^{3w} \log n)$ with $n$ processors, where $r$ is the maximum range of any variable, and $w$ is the induced width after a fill-in computation. Note that, for trees and polytrees, $w$ is a constant. The running time in multiply connected networks can be improved to $O(w \log n)$ if exponentially many $(r^{3w} n)$ processors are employed. Section 7 surveys some related work; section 8 summarizes, and addresses several open questions.

## 2  BAYESIAN NETWORKS

Consider a set of $n$ random variables, $W = \{A_1, A_2, \ldots, A_n\}$. The probability that variable $A_j$ takes on the value $a_j$ is $\Pr(A_j = a_j)$, where $a_j \in \{1, 2, \ldots, r\}$, and $r$ is a constant. A *joint probability distribution*, $\Pr(W) = \Pr(A_1, A_2, \ldots, A_n)$, assigns a probability to every possible combination of values of variables; it is a mapping from the cross-product of the ranges of the variables to the unit interval [0,1] (with a normalization constraint). The joint probability distribution thus consists of $r^n$ probabilities, an unmanageably large number even for relatively small $r$ and $n$. To simplify notation, $r$ is not explicitly indexed, even though it may vary across variables.

A *Bayesian network* exploits conditional independence to represent a joint distribution more compactly. Consider the variable $A_j$. Suppose that, given the values of a set of preceding variables $\mathbf{pa}(A_j)$—called $A_j$'s *parents*—$A_j$ is conditionally independent of all other preceding variables. Then we can rewrite the joint probability distribution in a (usually) more compact form:

$$\Pr(W) = \Pr(A_1, A_2, \ldots, A_n) = \prod_{j=1}^{n} \Pr(A_j | \mathbf{pa}(A_j)).$$

For each variable $A_j$, a *conditional probability table* (CPT) records the conditional probabilities $\Pr(A_j | \mathbf{pa}(A_j))$, for all possible combinations of values for $A_j$ and $\mathbf{pa}(A_j)$.

A Bayesian network can be represented graphically as a *directed acyclic graph* (DAG). Each variable is a node or vertex in the graph, and directed edges encode parent relationships. There is a directed edge from $A_i$ to $A_j$ if and only if $A_i$ is parent of $A_j$. We may also refer to $A_j$ as the child of $A_i$, and $\mathbf{ch}(A_i)$ as the set of children of $A_i$. We use $\mathbf{gp}(A_j) \equiv \mathbf{pa}(\mathbf{pa}(A_j))$ to denote the set of grandparents (parents of parents) of $A_j$. A DAG has no directed cycles and thus defines a partial order over its vertices. We assume without loss of generality that the variable indices are consistent with this partial ordering: if $A_i$ is an ancestor of $A_j$ then $i < j$. We denote a *total ordering* of the nodes, which may or may not be consistent with the DAG's partial order, as $\alpha = \{\alpha_1, \alpha_2, \ldots, \alpha_n\}$, where $\alpha$ is a permutation of $\{1, 2, \ldots, n\}$. Thus $A_{\alpha_3}$ is the third node according to ordering $\alpha$, etc.

A *polyforest* is a DAG with no undirected cycles. A *polytree* is a connected polyforest. A *forest* is a polyforest where every node has at most one parent. A *tree* is a connected forest. Since each tree in a forest can be handled independently, we need not explicitly consider forests. A *chain* is a tree where each node has at most one child. For complexity analysis, it is generally assumed that the maximum number of parents for any variable is constant, since the size of the input Bayesian network itself is exponential in the parent set size.[1] Given this assumption, standard serial algorithms for inference in trees and polytrees run in $O(n)$ time.

A set $C$ of nodes is called a *clique*, or a maximal complete set, if all of its nodes are fully connected, and no proper superset of $C$ is fully connected. Within a clique, inference takes $O(r^{|C|})$ time, since we must essentially sum over $C$'s entire joint distribution.

---

[1] This assumption breaks down for some more compact conditional probability encodings, not treated here, e.g., noisy-OR.

Most methods for Bayesian inference in multiply connected networks first convert to a (possibly exponentially larger) cycle-free representation, called variously a cluster tree, junction tree, clique tree, or join tree (Jensen, Lauritzen, and Olesen 1990; Lauritzen and Spiegelhalter 1988; Neapolitan 1990; Shachter, Andersen, and Szolovits 1994; Spiegelhalter, Dawid, Lauritzen, and Cowell 1993). First, we *moralize* the graph by fully connecting ("marrying") the parents of every node, and dropping edge directionality. Next, the graph is *filled in* according to a given total ordering of nodes $\alpha$: starting with $A_{\alpha_n}$, and continuing for $j = n - 1$ down to $j = 1$, all preceding neighbors of $A_{\alpha_j}$ (according to ordering $\alpha$) are fully connected. This fill-in computation ensures that the graph is *triangulated*, or has no four-node cycles without a chord. Each clique is then clustered into a *supernode* that can take on all possible combinations of values of its constituents. Supernode clusters, denoted $C_j$, are indexed according to their highest numbered constituent (according to $\alpha$). A triangulated graph has the *running intersection property*. For every $j$, there exists an $i < j$ such that:

$$C_j \cap (C_1 \cup C_2 \cup \cdots \cup C_{j-1}) \subseteq C_i$$

For every $j$, we choose one $i$ for which this property holds, and connect $C_j$ to $C_i$. The result is an undirected tree of cliques. In the final step, for every two adjacent cliques, their intersection (called a *separating set*) is clustered into a new supernode, and placed between the two. The *induced width* $w$ of a network, relative to an ordering $\alpha$, is defined as the maximum number of nodes in any clique, after moralization and fill-in. Computing the *optimal* $\alpha$, or equivalently the optimal triangulation, that yields the minimum induced width is itself an NP-complete problem (Klocks 1994)—for this reason, heuristic methods are usually employed. Serial algorithms for inference in cluster trees run in $O(r^w n)$ time.

If we now reintroduce directionality to the cluster tree edges, consistent with the *original* DAG partial order, the result is a polytree of supernodes; I call this a *directed cluster polytree*. Each clique's CPT is simply the product of its constituents' CPTs. Any polytree inference algorithm is directly applicable.

The general problem of Bayesian inference is to find the probability of each variable given a subset $E$ of evidence variables that have been instantiated with values. That is, compute $\Pr(A_j|E)$ for every $j$, where $E = \{A_{e_1} = a_{e_1}, A_{e_2} = a_{e_2}, \ldots, A_{e_c} = a_{e_c}\}$, and each $e_k \in \{1, 2, \ldots, n\}$.

## 3 AN EXAMPLE: FINDING MARGINALS IN CHAINS

Consider the problem of finding all marginal probabilities, $\Pr(A_j)$, in the chain network of Figure 2(a), using $n$ processors.

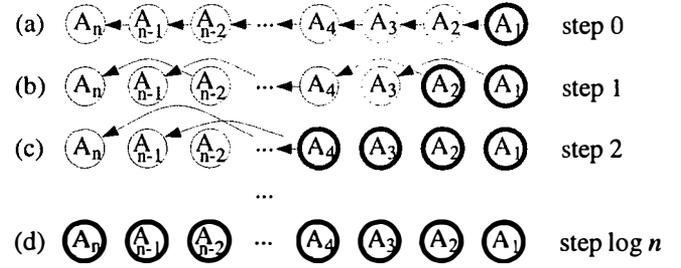

Figure 2: Propagating Probabilities in a Chain Network. At time step $t$, the first $2^t$ marginals, darkened in the figure, have been computed.

Processor $j$ is assigned to variable $A_j$, and "owns" the conditional probability table (CPT) for $\Pr(A_j|A_{j-1})$. Note that processor #1 already holds the marginal probability $\Pr(A_1)$; it simply records a flag indicating that it is done. At the first time step, each processor $j$ rewrites its CPT in terms of its *grandparent*, given its own current CPT and its parent's current CPT. Mathematically, processor $j$ computes:

$$\Pr(A_j|A_{j-2}) \leftarrow \sum_{A_{j-1}} \Pr(A_j|A_{j-1}) \Pr(A_{j-1}|A_{j-2}),$$

for each combination of values of $A_j$ and $A_{j-2}$, taking constant ($O(r^3)$) time. After time step one, processor #2 holds the marginal probability $\Pr(A_2)$, so it marks itself done (it knows that it is done because its parent was done at the previous time step). The new network topology after step one is pictured in Figure 2(b). The second time step proceeds exactly as the first—each variable rewrites its CPT in terms of its (new) grandparent:

$$\Pr(A_j|A_{j-4}) \leftarrow \sum_{A_{j-2}} \Pr(A_j|A_{j-2}) \Pr(A_{j-2}|A_{j-4}).$$

After time step two, the first four processors hold their respective marginals and mark themselves done. After time step three, the first eight processors are done, etc. After $\log_2 n$ time steps, all $n$ processors are done, and all marginal probabilities have been computed; the entire process is conveyed in Figure 2. This procedure of iteratively recomputing CPTs in terms of grandparents is analogous to a widely used method in parallel algorithms to achieve $O(\log n)$ computations in linked lists, called *pointer jumping*. In chain networks, the algorithm can operate in an exclusive-read fashion, though we will use concurrent reads later, to process trees and polytrees.

## 4 INFERENCE IN TREES

This section generalizes PHISHFOOD to compute inference queries in tree networks. The next subsection begins with computing marginal probabilities only; Section 4.2 presents a parallel arc reversal procedure to propagate evidence.

PHISH-TREE-MARGINALS($T$)
  INPUT:   Bayesian network tree $T$
  OUTPUT: $\Pr(A_j)$ for all $j$

1. mark root node $A_1$ done
2. **while** there is a node not marked done
3.    **for** each node $A_j$ **in parallel**
4.       compute (1)
5.       **if** $\text{pa}(A_j)$ is marked done
6.          **then** mark $A_j$ done
7.          **else** $\text{pa}(A_j) \leftarrow \text{gp}(A_j)$

Table 1: Computing all Marginal Probabilities in a Tree Network.

### 4.1 FINDING MARGINALS IN TREES

Marginal computation in trees proceeds in much the same way as for chains. Again, the central step is for each variable to rewrites its CPT in terms of its grandparent:

$$\Pr[A_j|\text{gp}(A_j)] \leftarrow \qquad (1)$$
$$\sum_{\text{pa}(A_j)} \Pr[A_j|\text{pa}(A_j)] \cdot \Pr[\text{pa}(A_j)|\text{gp}(A_j)].$$

Because the graph is a tree, each "set" $\text{pa}(A_j)$ and $\text{gp}(A_j)$ is actually a singleton. The procedure psuedocode is given in Table 1. This simple implementation uses concurrent reads, since each variable may have several grandchildren at a given step. If we consider the root to be at depth one, then, after time step two, all marginals for variables at depth two have been computed; after step three, all marginals at depth four have been computed, etc. All marginals in the tree are computed in $O(\log d)$ time, where $d$ is its depth and, in the worst-case, is $O(n)$. Notice that, if the tree is balanced (i.e., is of depth $O(\log n)$), then the running time is $O(\log \log n)$.

### 4.2 EVIDENCE PROPOGATION IN TREES

This subsection describes how PHISHFOOD propogates evidence, using a parallel version of *arc reversal* (a form of Bayes's rule), and *evidence absorption*, the standard versions of which were developed by Shachter (1988, 1990).

We assume that the graph topology is given as an adjacency list, so that each node has pointers to its children as well as to its parent. We utilize a standard parallel subroutine that finds a *preorder walk* of a tree, starting from any given root node. A preordering numbers the nodes according to the order that they are first reached in a depth-first traversal of the underlying undirected tree. The computation can be done in $O(\log n)$ time with $n$ processors, using what is called the *Euler tour technique* (Gibbons and Spirakis 1993; JáJá 1992; Tarjan and Vishkin 1985).

In Bayesian networks, any edge can be reversed such that the transformed network represents the same underlying joint distribution, as long the reversal does not create a directed cycle. When an edge between $A_i$ and $A_j$ is reversed, both variables must share each other's parents, which, in general, may add significantly to the number of arcs in the graph. However, a tree can be rerooted at any node without adding any arcs. Consider rerooting the tree at the first evidence variable, $A_{e_1}$. Let $\alpha$ be a preorder walk of the underlying undirected tree, starting from $A_{e_1}$. Then $\alpha$ encodes the desired new edge directions—that is, an edge in the rerooted tree points from $A_{\alpha_i}$ to a neighbor $A_{\alpha_j}$ if and only if $\alpha_i < \alpha_j$. A serial arc reversal procedure would begin at the original root $A_1$, which has no parent, reversing any of $A_1$'s child arcs inconsistent with $\alpha$. $A_1$'s children have no parents other than $A_1$ itself, and thus no new arcs are created. Next, any child arcs from $\text{ch}(A_1)$ inconsistent with $\alpha$ are reversed, then any child arcs from $\text{ch}(\text{ch}(A_1))$, etc. The result is a new tree, rooted at $A_{e_1}$, with the same underlying undirected topology as the original tree.

Suppose that all marginal probabilities are known, computed as in Section 4.1. Then, since the tree topology is invariant, it is not necessary to reverse arcs in any particular order. We can simply reverse any edges inconsistent with $\alpha$ simultaneously, in constant parallel time, by applying Bayes's rule. To be concrete, assign one processor to every edge (since the graph is a tree, there are $n - 1$ edges). If an arc from $A_i$ to $A_j$ is inconsistent with the ordering $\alpha$, then the processor reads the CPT associated with $A_j$ and rewrites the CPT associated with $A_i$ according to Bayes's rule:

$$\Pr(A_i|A_j) \leftarrow \frac{\Pr(A_j|A_i)\Pr(A_i)}{\Pr(A_j)}. \qquad (2)$$

Since the original and new structures are both trees, each processor reads from and writes over at most one CPT, and each CPT is rewritten by at most one processor.

In general, *evidence absorption* (Shachter 1988; Shachter 1990) involves a single update at each of $A_{e_1}$'s children, and a series of arc reversals propagating to all of $A_{e_1}$'s ancestors. However, by construction, the evidence variable is the root and has no ancestors. Each of $A_{e_1}$'s children $A_j$ simply absorbs the evidence in constant time, with the update:

$$\Pr(A_j) \leftarrow \Pr(A_j|A_{e_1} = a_{e_1}). \qquad (3)$$

Next, all of $A_{e_1}$'s child arcs are eliminated, creating disconnected trees rooted at each of $A_{e_1}$'s children. The evidence propagation and absorption process is depicted in Figure 3. The implicit marginal probabilities in these new trees are actually posterior probabilities, given the evidence $A_{e_1} = a_{e_1}$, in the original representation. Note that evidence given in the form of a likelihood function could also be absorbed at $A_{e_1}$; in this case, $A_{e_1}$'s child arcs would in general not be eliminated. The algorithm repeats by computing the new marginals as described in Section 4.1, rerooting the tree at $A_{e_2}$, and absorbing the second evidence

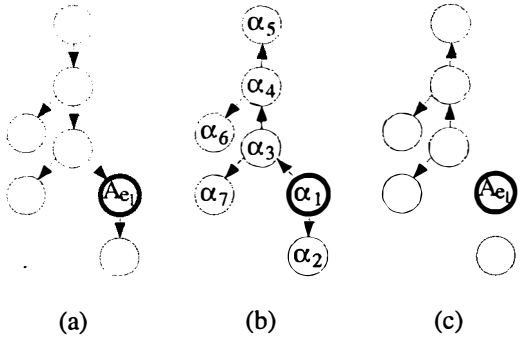

Figure 3: Propagating Evidence in a Tree Network. (a) The initial tree. (b) A preordering $\alpha$ of nodes, starting from $A_{e_1}$, with inconsistent arcs reversed. (c) Evidence at the new root, $A_{e_1}$, is absorbed.

PHISHFOOD-TREE($T, E$)
  INPUT:  Bayesian network tree $T$, adjacency list,
          and $c$ evidence variables and values:
          $E = \{A_{e_1} = a_{e_1}, \ldots, A_{e_c} = a_{e_c}\}$
  OUTPUT: $\Pr(A_j|E)$ for all $j$

1. PHISH-TREE-MARGINALS($T$)
2. **for** each evidence variable $A_{e_k}$
3.    $\alpha \leftarrow$ PREORDER($T, A_{e_k}$)
4.    **for** each edge $(A_i, A_j)$ **in parallel**
5.       **if** $(A_i, A_j)$ is inconsistent with $\alpha$ **then**
6.          compute (2) and store in $A_i$'s CPT
7.    absorb evidence (3) at new root $A_{e_k}$
8.    PHISH-TREE-MARGINALS($T$)

Table 2: Parallel Bayesian Inference in a Tree Network.

variable, etc. The full psuedocode is given in Table 2. For any constant number of evidence variables, its running time is $O(\log n)$ with $n$ processors.

## 5 INFERENCE IN POLYTREES

This section describes a parallel algorithm for Bayesian inference in polytree networks. Once again, discussion is divided into two stages: (1) computing marginal probabilities, and (2) propogating evidence.

### 5.1 FINDING MARGINALS IN POLYTREES

In trees, PHISHFOOD's key to parallelization is for each variable to iteratively rewrite its CPT in terms of its grandparent. In polytrees, a node may have more than one parent, each of which has more than one parent, etc. Since CPT size is exponential in the number of parents, a simple application of PHISH-TREE-MARGINALS would be worst-case intractable, so a modified approach is necessary.

The structure of a polytree implies many independencies between variables, which can be exploited to speed inference (Neapolitan 1990; Pearl 1988; Russell and Norvig 1995). In particular, given no evidence at $A_j$, or at any of $A_j$'s descendents, the parents of $A_j$ are independent. That is, if $A_{i_1}$ and $A_{i_2}$ are both parents of $A_j$, then $\Pr(A_{i_1} A_{i_2}|E) = \Pr(A_{i_1}|E)\Pr(A_{i_2}|E)$, for any set of evidence variables $E$ that does not contain $A_j$ or any of $A_j$'s descendents. Consider an arbitrary node $A_j$. If its parents' marginal probabilities are known, then, since its parents are independent, $A_j$'s marginal can be computed. Similarly, its parents' marginals can be computed once its grandparents' marginals are known. Let $A_j$'s *induced ancestral polytree*[2] (IAP) be the subnetwork induced by $A_j$ and its ancestors—that is, the network consisting of $A_j$, $\mathbf{pa}(A_j)$, $\mathbf{pa}(\mathbf{pa}(A_j))$, $\mathbf{pa}(\mathbf{pa}(\mathbf{pa}(A_j)))$, etc., and the edges between them. Note that this subnetwork contains all of the necessary information to compute $\Pr(A_j)$.

I will make use of the following terms: a *polynode* is a variable with more than one parent, a *treenode* is a variable with exactly one parent, and a *rootnode* is a variable with no parents.

I adopt a strategy similar in style to that employed in Miller and Reif's *parallel tree contraction* algorithm (1985). The algorithm consists of two main steps: the first is called *rootnode absorption* (analogous to Miller and Reif's RAKE procedure), and the second is *treenode jumping* (analogous to their COMPRESS procedure). The main conceptual difficulty in directly mapping the current problem to parallel tree contraction is that each node can have both multiple parent and multiple child relationships, each with very different semantics. The key insight is that each node's IAP can be contracted in more or less the standard way, by exploiting the independence properties mentioned above.

We initialize the algorithm by marking all rootnodes as done. In the first step, each node absorbs all of its rootnode parents (if any) in constant parallel time. Consider node $A_j$. Denote $R_j = \{A_{i_1}, \ldots, A_{i_q}\}$ as the set of $A_j$'s rootnode parents, and $I_j$ as the set of $A_j$'s other, internal node parents (thus $\mathbf{pa}(A_j) = R_j \cup I_j$). Then the update at $A_j$ is as follows:

$$\Pr(A_j|I_j) \leftarrow \sum_{R_j} \Pr(A_j|R_j \cap I_j)\Pr(R_j|I_j) \qquad (4)$$
$$= \sum_{R_j} \Pr(A_j|R_j \cap I_j)\Pr(A_{i_1})\cdots\Pr(A_{i_q})$$

where $\sum_{R_j}$ is the sum over all combinations of values of the variables in $R_j$, and the equality follows from the independence properties of a polytree. Any new rootnodes that are formed after absorption now hold their marginal probabilities, and are marked as done. This first phase is illustrated in Figure 4.

---
[2]Indiscriminate use of this phrase discouraged.

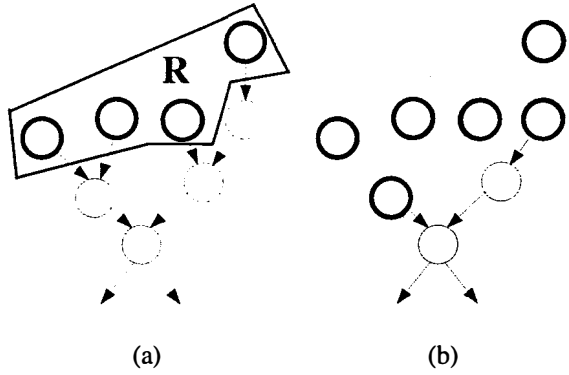

(a)                       (b)

Figure 4: Rootnode Absorption.

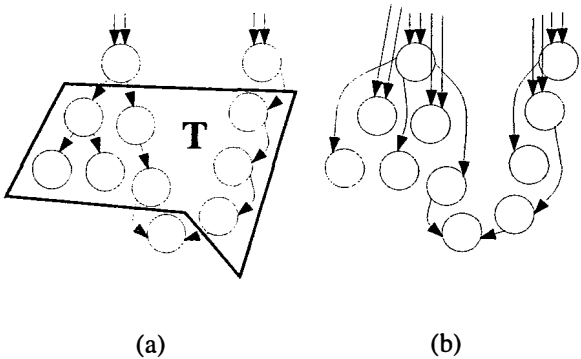

(a)                       (b)

Figure 5: Treenode Jumping.

Iterative application of rootnode absorption alone would require $O(n)$ time to compute all marginals in an unbalanced polytree. We interleave the second step—treenode jumping—to compress long treenode chains in the network. We accomplish this with one step of the pointer jumping procedure described in section 3, applied only at treenodes. That is, each treenode CPT is rewritten in terms of its grandparent(s) by computing (1). Since treenodes have exactly one parent, this step does not introduce a combinatorial growth in parent set size. This step will essentially reduce the length of treenode chains in each IAP by half. As before, if the treenode's parent is marked done at time step $t-1$, then the treenode itself can be marked done at time step $t$. This second phase is pictured in Figure 5. After both phases, the algorithm iterates on the transformed polytree: we absorb each new rootnode, rewrite each new treenode CPT in terms of its grandparent(s), and repeat until all nodes are marked done. The algorithm psuedocode is given in Table 3.

For the time complexity analysis, we show that, for each node, the number of "active" nodes (those not marked done) in its induced ancestral polytree (IAP) is reduced by at least half at each iteration. Consider $A_j$'s IAP at time step $t-1$, consisting of rootnodes $R$, treenodes $T$, and polynodes $P$. Note that because the network is acyclic, $|P| \leq |R|$. (Each

```
PHISH-POLY-MARGINALS(P)
  INPUT:  Bayesian network polytree P
  OUTPUT: Pr(A_j) for all j

1.  mark all rootnodes done
2.  while there is a node not marked done
3.    for each node A_j with rootnode parent(s)
      in parallel
4.      absorb rootnode parents (4)
5.      if A_j has no more parents
6.        then mark A_j done
7.    for each treenode A_j in parallel
8.      compute (1)
9.      if pa(A_j) is marked done
10.       then mark A_j done
11.       else pa(A_j) ← gp(A_j)
12.   identify new rootnodes, treenodes, polynodes
```

Table 3: Computing all Marginal Probabilities a Polytree Network.

polynode implies a path between at least two rootnodes; if there are more polynodes than rootnodes, then there must be a cycle in the IAP.) At time step $t$, $A_j$'s new IAP consists of the nodes $T' \cup P'$, where $T' \subseteq T$, and $P' \subseteq P$, since all rootnodes have been absorbed. For each node in $T'$, exactly one node in $T \cup P$ has been "jumped", and is thus no longer in $A_j$'s IAP. Thus, $|T'| = |T| - |T'| + |P| - |P'|$. Combining this equality with the previous inequality, we find that:

$$|T'| + |P'| \leq |R| + |T| - |T'| + |P| - |P'|$$
$$|T'| + |P'| \leq \frac{1}{2}(|R| + |T| + |P|).$$

Thus, after $O(\log n)$ iterations, each node's IAP is reduced to a singleton, and all marginals have been computed.

### 5.2 EVIDENCE PROPAGATION IN POLYTREES

As with trees, evidence is propagated with arc reversals. The first step is to convert the polytree into a directed cluster polytree, as described in Section 2. Since the moral graph of a polytree is triangulated, the supernodes simply consist of each variable and its parents, and the cluster polytree's size is of the same order as the original polytree. A cluster tree, like an ordinary tree, can be rooted at any node, without changing its underlying undirected topology (Chyu 1991; Shachter, Andersen, and Poh 1991). Once all marginal probabilities are known, we can root the cluster tree with the method described in Section 4.2 in $O(\log n)$ time. The algorithm PHISHFOOD-TREE is then applicable, unmodified, for inference in the rooted cluster tree.

## 6  INFERENCE IN ARBITRARY NETWORKS

In the general case, we first convert to a directed cluster polytree, as described in Section 2. The parallel inference algorithm, called McPHISHFOOD (Multiply Connected PHISHFOOD) is exactly that for polytrees, and has a new acronym only to accommodate any future corporate sponsorship.

Assuming that the fill-in ordering $\alpha$ is chosen with a heuristic method, all of the graph-theoretic computations can be done in polynomial time. Consider the complexity of computing the supernodes' CPTs. After separation sets are introduced, each supernode CPT contains $O(r^w)$ entries, and each entry is computed by multiplying $w$ of the original CPTs' entries. For this task, parallelization is trivial. Each entry of each supernode CPT can be computed independently, taking $O(r^w w)$ time with one processor per clique, or $O(\log w)$ time with $r^w w$ processors per clique. [3]

In both the tree and polytree algorithms, the complexity bottleneck is (1), because this computation involves three supernodes. The summation is evaluated for each possible combination of values of $C_j$ and $\text{gp}(C_j)$. A conservative bound for computing the new CPT is $O(r^{3w})$ time with one processor, or $O(\log r^w) = O(w)$ time with $r^{3w}$ processors. We can do slightly better by exploiting separation sets. First, each separation set CPT is rewritten in terms of its separation set grandparent(s), taking $O(r^w)$ time with one processor, or $O(1)$ time with $r^w$ processors. The marginal-finding algorithm is then applied only to the network of separation sets. Once all separation set marginals are known, remaining marginals can be computed in $O(r^w)$ or $O(1)$ time, with one or $r^w$ processors, respectively.

Any cluster tree can be rooted at any of its cliques, without changing its topology (Chyu 1991; Shachter, Andersen, and Poh 1991). For each evidence variable, we root the cluster tree at the supernode that contains it. Each application of Bayes's rule (2) takes $O(r^w)$ time with one processor, or $O(1)$ time with $r^w$ processors. Evidence is absorbed at the new root by instantiating the variable, then renormalizing the remaining constituents of the supernode, in $O(r^w)$ or $O(w)$ time, with one or $r^w$ processors, respectively.

The full McPHISHFOOD algorithm then runs in $O(r^{3w} \log n)$ time with $n$ processors, or $O(w \log n)$ time with $r^{3w} n$ processors, for any constant number of evidence variables. Let $s$ be the maximum separation set size. If we exploit separation sets as described above, then inference takes $O(r^{3s} \log n)$ time with one processor, or $O(w + s \log n)$ time with $r^{3s} n$ processors. These improved bounds assume that $3s > w$.

---

[3]Multiplication (or addition) of $m$ numbers can be done in $O(\log m)$ parallel time with $m$ processors.

## 7  RELATED WORK

D'Ambrosio (1992) examines the possibility of parallelizing the *symbolic probabilistic inference* (SPI) algorithm on hypercube parallel architectures. He identifies two sources of concurrency: topological or "evaluation tree" parallelism, and conformal product parallelism; in the former case, he only considers exploiting independent computations, stating that, "a lower bound on the running time of an algorithm exploiting evaluation tree parallelism can be calculated by summing the [computations] in the longest path of the evaluation tree." Given several different modeling assumptions, he empirically evaluates predicted computational, communication, and storage costs, and parallel speedup and efficiency. He concludes that good opportunities exist for parallelizing independent conformal products evaluations, but not much natural concurrency is present at the topological level.

Kozlov and Singh (1996, 1994) describe experimental results for a parallel version of a junction tree algorithm, implemented on a Stanford DASH multi-processor and an SGI Challenge XL. They identify the same two types of concurrency (topological and in-clique), and also only considered parallelizing independent operations: "computations for cliques that do not have an ancestor-descendent relationship can be done in parallel." (Kozlov 1996).

Delcher et al. (1996) give an $O(\log n)$ time *serial* algorithm for inference queries in tree networks, after a linear time preprocessing step to generate a special data structure. This data structure is actually based on the tree contraction sequence described by Abrahamson et al. in the context of a parallel algorithm (1989).

## 8  CONCLUSION AND FUTURE WORK

I have presented PHISHFOOD, a parallel algorithm for exact Bayesian inference that, for any constant number of evidence variables, runs in $O(\log n)$ time for polytrees, and $O(r^{3w} \log n)$ time for cluster trees with $n$ processors, and $O(w \log n)$ time for cluster trees with $r^{3w} n$ processors. Concurrent memory reads, but not writes, are required.

There are several possible directions for improving the algorithm. An $O(\log n)$ exclusive-read polytree algorithm may be possible, perhaps based on Abrahamson et al.'s EREW parallel tree contraction algorithm (1989).[4] The *work* done by an algorithm is defined as its running time multiplied by the number of processors used. There may exist a work-efficient version of PHISHFOOD that uses only $n/\log n$ processors.

Future work might also pursue parallelizations for other

---

[4]Any CREW (or CRCW) algorithm can run on an EREW PRAM with an $O(\log n)$ factor slowdown. (Cormen, Leiserson, and Rivest 1992; Gibbons and Spirakis 1993; JáJá 1992)

types of inference queries (e.g., most probable explanation, maximum expected utility, value of information), and for other conditional probability encodings (e.g., noisy-OR).

Certainly, an implementation of these algorithms is called for, on a specific parallel architecture, along with empirical comparisons with existing serial and parallel algorithms.

**Acknowledgments**

Thanks to David Pynadath and to the anonymous reviewers.